\documentclass[runningheads]{llncs}
\usepackage[T1]{fontenc}
\usepackage{amsmath}
\usepackage{amssymb}
\usepackage{xspace}
\usepackage{graphicx}
\usepackage{multirow}
\usepackage{color, colortbl}

\definecolor{LightCyan}{rgb}{0.88,1,1}
\usepackage[pagebackref,breaklinks,colorlinks,citecolor=blue]{hyperref}
\usepackage[capitalize]{cleveref}
\crefname{section}{Sec.}{Secs.}
\Crefname{section}{Section}{Sections}
\Crefname{table}{Table}{Tables}
\crefname{table}{Table}{Tables}

\newcommand{\C}[1]{\mathcal{#1}}
\newcommand{\bd}[1]{\boldsymbol{#1}}
\newcommand{\best}[1]{\textcolor{red}{\underline{#1}}}
\newcommand{\sbest}[1]{\textcolor{blue}{\textit{#1}}}
\makeatletter
\DeclareRobustCommand\onedot{\futurelet\@let@token\@onedot}
\def\@onedot{\ifx\@let@token.\else.\null\fi\xspace}

\def\ie{\emph{i.e}\onedot}

\makeatother
\begin{document}
\title{TAKT: Target-Aware Knowledge Transfer for Whole Slide Image Classification}
\titlerunning{TAKT for WSI Classification}
\author{Conghao Xiong\inst{1}\thanks{Equal contribution.}\thanks{Corresponding authors: \email{chxiong21@cse.cuhk.edu.hk, howzheng@tencent.com}.} 
\and Yi~Lin\inst{2}$^\dagger$ 
\and Hao~Chen\inst{2,3} 
\and Hao~Zheng\inst{4}$^\ddagger$ 
\and Dong~Wei\inst{4} 
\and Yefeng~Zheng\inst{4} 
\and Joseph~J.~Y.~Sung\inst{5} 
\and Irwin~King\inst{1}} 
\authorrunning{C. Xiong et al.}
%
\institute{Dept.~of Computer Science and Engineering, The Chinese University of Hong Kong \and
Dept.~of Computer Science and Engineering, The Hong Kong University of Science and Technology (HKUST) \and
Dept.~of Chemical and Biological Engineering, HKUST \and 
Jarvis Research Center, Tencent YouTu Lab \and
Lee Kong Chian School of Medicine, Nanyang Technological University
}
%
%
\maketitle              
\begin{abstract}
Transferring knowledge from a source domain to a target domain can be crucial for whole slide image classification, since the number of samples in a dataset is often limited due to high annotation costs. However, domain shift and task discrepancy between datasets can hinder effective knowledge transfer. In this paper, we propose a Target-Aware Knowledge Transfer framework, employing a teacher-student paradigm. Our framework enables the teacher model to learn common knowledge from the source and target domains by actively incorporating unlabelled target images into the training of the teacher model. The teacher bag features are subsequently adapted to supervise the training of the student model on the target domain. Despite incorporating the target features during training, the teacher model tends to overlook them under the inherent domain shift and task discrepancy. To alleviate this, we introduce a target-aware feature alignment module to establish a transferable latent relationship between the source and target features by solving the optimal transport problem. Experimental results show that models employing knowledge transfer outperform those trained from scratch, and our method achieves state-of-the-art performance among other knowledge transfer methods on various datasets, including TCGA-RCC, TCGA-NSCLC, and Camelyon16. Codes are available at \url{https://github.com/BearCleverProud/TAKT}.

\end{abstract}
\section{Introduction}
\label{sec:intro}

\textbf{W}hole \textbf{S}lide \textbf{I}mage (WSI) refers to the digitised glass slides containing histology tissues, which is crucial for cancer diagnosis. Consequently, WSI classification has become a research focus, including skin \cite{ianni_tailored_2020}, lung \cite{chen_annotation-free_2021}, breast \cite{litjens_deep_2016}, prostate \cite{campanella_clinical-grade_2019}, and pancreas \cite{keikhosravi_non-disruptive_2020} cancers, while its success heavily depends on the availability of a large set of labeled samples. However, annotating WSIs is labour-intensive, and in certain cases, the patient cohort is limited, resulting in a restricted number of WSIs in one dataset \cite{amupan_predictors_2021}, limiting the data-hungry deep learning techniques.


An effective solution is to acquire knowledge from other datasets. Malignant cells share similar morphological characteristics, such as enlarged nuclei, irregular size and shape, prominent nucleoli, and intense or pale cytoplasm \cite{baba2007comparative}. The shared characteristics can be leveraged to alleviate the over-fitting problem in relatively small datasets \cite{pan_survey_2010,yu_transfer_2022,zhuang_comprehensive_2021}. However, in practice, knowledge transfer could suffer from two challenges, domain shift and task discrepancy. 

Domain shift is a change in distribution, caused by differences in tumour sizes and colour tones across WSIs. Models trained on one dataset are often biased towards it, limiting knowledge transfer. Task discrepancy is inconsistency in tasks between domains, and task-specific features may not directly transfer to another task. Previous works address domain shift from the domain adaptation perspective \cite{ahn_2020_uda,feng2021deep}. However, they focus on style differences and assume that the label space remains the same, while the label spaces are different in our case.

To tackle these issues, we propose a \textbf{T}arget-\textbf{A}ware \textbf{K}nowledge \textbf{T}ransfer (TAKT) framework to enable the teacher model to learn common knowledge across two domains in a teacher-student paradigm. Within this framework, we actively incorporate unlabelled data from the target domain during the training of the teacher model, using our proposed \textbf{T}arget-\textbf{A}ware \textbf{D}ata \textbf{A}ugmentation (TADA) method. When augmenting a source feature, we retrieve the closest centroids generated by an unsupervised clustering algorithm from the target dataset and integrate them into the source features. These centroids are the most representative and relevant samples from the target domain, which enhance the diversity of the augmented dataset without compromising its original distribution. This approach has two rationales. Firstly, integrating target domain representations enhances the transferability of learned knowledge in a specified direction, \ie, from the source domain to the target domain. Secondly, rich data augmentation inherently helps the model learn more generalised features.

However, it is observed that due to the domain shift and task discrepancy, the teacher model still tend to overlook the target features after TADA, and hence the teacher model remains biased towards the source domain. To address this issue, we introduce a \textbf{T}arget-\textbf{A}ware \textbf{F}eature \textbf{A}lignment (TAFA) module, applied between features from source and target domains. This module establishes a transferable relationship between the source and target features by solving an \textbf{O}ptimal \textbf{T}ransport (OT) problem, providing a holistic view of the two distributions. This relationship enforces the teacher model to pay similar attention to the target features, enabling the model to learn more common knowledge across both domains. The contributions of this paper are summarised below: 
\begin{enumerate}
    \item We propose a TAKT framework to alleviate the impact of domain shift by actively incorporating unlabelled target data with the source domain.
    \item We propose a TAFA module to mitigate bias towards the source domain by establishing a latent relationship between source and target features.
    \item We conduct extensive experiments on three datasets, including RCC, NSCLC and Camelyon16. Results show the efficacy of proposed method.
\end{enumerate}

\begin{figure*}[t]
    \centering
    \includegraphics[width=\linewidth]{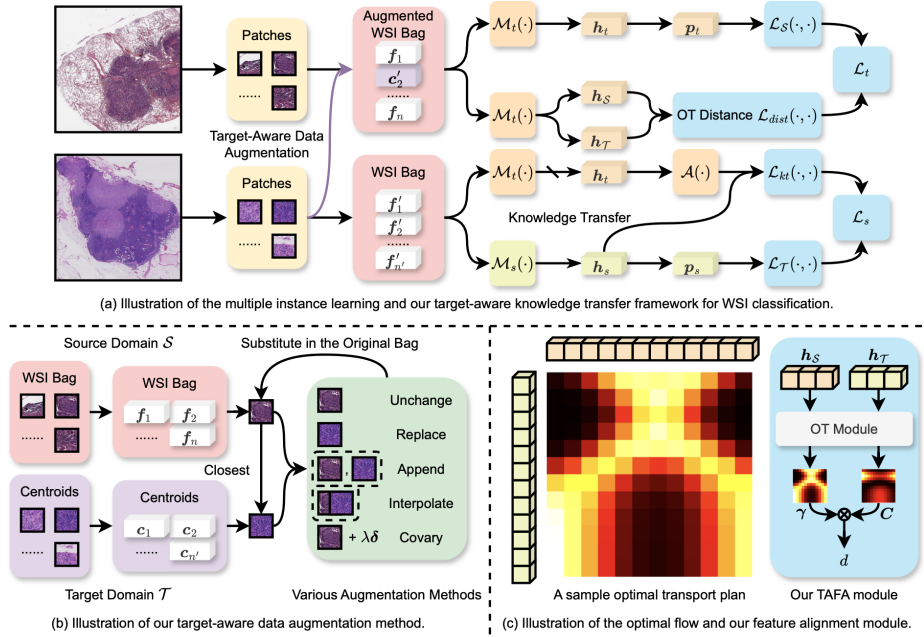}
    \caption{Illustrations of (a) our target-aware knowledge transfer framework, (b) our target-aware data augmentation method and (c) our target-aware feature alignment module and a sample optimal transport flow. $\C{M}_{t}(\cdot)$ is the teacher model and $\C{M}_{s}(\cdot)$ is the student model. The light areas in (c) indicate regions with higher values.}
    \label{fig:framework}
\end{figure*}
\section{Methodology}


\subsection{Overview of the Target-Aware Knowledge Transfer}
The TAKT framework, shown in \cref{fig:framework}(a), has two parts: a TADA method and a TAFA module. Following the teacher-student paradigm, our framework involves training a teacher model on a dataset generated with TADA. The student model is trained on the target dataset, supervised by the adapted teacher bag features. 

Mathematically, the loss function for the student model $\C{L}_s$ is given as, 
\begin{equation}\label{eq:our_loss}
    \C{L}_s = \C{L}_{\C{T}} + \alpha \C{L}_{kt} = \C{L}_{\C{T}} + \alpha \sum\left(\operatorname{MHA}\left(\operatorname{sign}(\bd{h}_t)|\frac{\bd{h}_t}{T}|^{\frac{1}{n}}\right) - \bd{h}_{s}\right)^2,
\end{equation}
where $\C{L}_{\C{T}}$ is the loss function on the target domain $\C{T}$, $\alpha$ is a coefficient, $\operatorname{MHA}(\cdot)$ is the \textbf{M}ulti-\textbf{H}ead \textbf{A}ttention (MHA) \cite{transformer_2017}, used to adapt source-specific features from the teacher model to target-specific ones that are more suitable for the student model to learn, $\bd{h}_{t}, \bd{h}_{s}$ are the teacher and student bag features, respectively, and $T=0.1, n=3$ are coefficients in \textbf{P}ower \textbf{T}emperature \textbf{S}caling \cite{he_knowledge_2022}.


\subsection{Target-Aware Data Augmentation}
\subsubsection{Overview of the TADA Method.} The illustration of TADA is shown in \cref{fig:framework}(b). The augmentation is performed at the instance level between WSI bags, while for target WSI bags, clustered centroids instead of raw patch features are leveraged. During the augmentation, we find the closest centroid for each source feature. The augmentation operations are designed based on ReMix \cite{yang_remix_2022}.

\subsubsection{Step 1: Clustering of the Target Dataset.} Each WSI bag from the target domain  $\C{T}$ is clustered using $K$-means \cite{lloyd_kmeans_1982}, which discovers highly representative features within that bag. Consequently, the feature vectors are transformed into more representative centroids that can effectively represent a set of features. The source dataset remains unchanged due to the potential adverse effects of clustering on the dataset, such as a significant reduction in the number of instances, in which case, augmentation techniques would greatly change the original distribution of the source dataset, thereby negatively affecting the performance of the teacher model and its transferability to downstream tasks.

\subsubsection{Step 2: Cross-Dataset Mix Operation.} 
The motivation behind our TADA is to enrich the source features by imparting them with a specific ``direction'' to transfer, \ie, towards the target domain. Trained with the target data, the teacher model grasps the general target feature distribution, enabling it to learn source features that exhibit better generalisability compared to the teachers that have not been exposed to the target data. Specifically, for each WSI bag in the source dataset, we enumerate each feature and augment the feature with a fixed probability $p$. For each feature $\bd{f}$ selected for augmentation, we first identify the closest centroid vector $\bd{c}$ from the stacked centroid matrix of the target domain. With the closest centroid vector $\bd{c}$, we employ the following operations:
\begin{enumerate}
    \item\textbf{Append.} $\bd{c}$ is appended to the bag. 
    \item\textbf{Replace.} $\bd{f}$ is replaced with $\bd{c}$ in the bag.
    \item\textbf{Interpolate.} The interpolation vector $\bd{f}_{I} = (1-\lambda)\bd{f} + \lambda \bd{c}$ is appended to the bag, where $\lambda \in (0,1)$ is the strength of the augmentation. 
    \item\textbf{Covary.} The vector $\bd{f}_{C} = \bd{f} + \lambda \bd{\delta}$ is appended to the bag. $\bd{\delta} \sim \C{N}(\bd{0}, \Sigma_{\bd{c}})$, where $\Sigma_{\bd{c}}$ is the covariance matrix of $\bd{c}$ and $\C{N}(\cdot,\cdot)$ is the normal distribution. 
    \item\textbf{Joint.} All the aforementioned methods are applied to the bag.
\end{enumerate}

\subsection{Target-Aware Feature Alignment Module}
Most of the recent methods for WSI fall under the category of attention-based multiple instance learning \cite{ilse_attention-based_2018,lu_data-efficient_2021,shao_transmil,xiong_diagnose_2023}. These methods calculate attention scores based on patch features using a gated attention network \cite{ilse_attention-based_2018} to determine the importance of each patch in the final prediction. In our framework, we observed that attention scores for target features are often lower than those for the source domain, indicating the teacher model is still biased towards the source domain after TADA. This bias hampers the effectiveness of data augmentation and diminishes the transferability of teacher features. However, directly penalising the low attention scores is not reasonable since target features may not necessarily contain source-specific features. Therefore, we propose a target-aware feature alignment module to establish a transferable latent relationship between source and target features instead of pursuing complete consistency, and this also prevent the target features to be neglected during training. 



As shown in \cref{fig:framework}, the inputs to TAFA module are source features and their closest target centroids. Within the TAFA module, how to establish the transferable latent relationship is treated as an OT problem. The OT distance is derived from the transportation plan with the lowest cost, providing a comprehensive view of the difference between two distributions in terms of shape, density, and spread, which are lacking in the traditional measures, such as Euclidean distance and cosine similarity.  
Mathematically, the input to the teacher model is denoted as $\bd{F} = [\bd{F}_{\C{S}}, \bd{C}_{\C{T}}]$, where $\bd{F}_{\C{S}} , \bd{C}_{\C{T}}$ are the feature vectors and centroids from the source and target domains, respectively. The bag features from the teacher model only using $\bd{F}_{\C{S}}$ and $\bd{C}_{\C{T}}$ are denoted as $\bd{h}_{\C{S}}$ and $\bd{h}_{\C{T}}$, respectively. For each input pair $(\bd{h}_{\C{S}}, \bd{h}_{\C{T}})$, we solve for the unbalanced OT problem,
\begin{equation}\label{eq:unbalanced_ot}
    \min_{\bd{\gamma}}<\bd{\gamma}, \bd{C}>_F + r\cdot\operatorname{\Omega}(\bd{\gamma}) + r_1\cdot\operatorname{KL}(\bd{\gamma}\bd{1}, \bd{a}) + r_2\cdot\operatorname{KL}(\bd{\gamma}^T\bd{1}, \bd{b}),
\end{equation}
where the ($i$, $j$)-th element in the cost matrix $\bd{C}$ is $|\bd{h}_{{\C{S}}_i} -\bd{h}_{{\C{T}}_j}|$, $<\cdot, \cdot>_F$ is the Frobenius inner product, $\operatorname{\Omega}(\bd{\gamma}) = \operatorname{KL}(\bd{\gamma}, \bd{a}\bd{b}^T)$ is the entropic regularisation term, $r, r_1, r_2$ are the regularisation coefficients, $\bd{a},\bd{b} = [1/c,\cdots,1/c]$ are two uniform distributions of size $c$, $c$ is the dimension of $\bd{h}_{\C{S}}$ and $\bd{h}_{\C{T}}$, $\operatorname{KL}(\cdot, \cdot)$ is the Kullback–Leibler divergence, and $\bd{\gamma} \ge 0$ is the optimal transport flow. 
The optimal flow indicates how teacher features can be transformed to have the same distribution as the student feature at minimal cost, considering the amount of mass transferred. The calculated distance between these two features decreases as the mass moved decreases.
The problem can be solved using the Sinkhorn-Knopp algorithm \cite{frogner_learning_2015}. Having obtained the optimal flow $\bd{\gamma}$, we calculate the distance $d = \sum\bd{C}\bd{\gamma}$. The overall loss function of the teacher model $\C{L}_t$ is given as,
\begin{equation}\label{eq:reg}
    \C{L}_t = \C{L}_{\C{S}} + \beta \C{L}_{dist} = \C{L}_{\C{S}} +\beta \sum\bd{C}(\bd{h}_{\C{S}}, \bd{h}_{\C{T}})\cdot\bd{\gamma}(\bd{h}_{\C{S}}, \bd{h}_{\C{T}}),
\end{equation}
where $\C{L}_{\C{S}}$ is the task-specific loss function on the source domain, and $\beta > 0$ is the coefficient. 
Within this formulation, the latent relationship is represented as the optimal flow, and the optimisation of the regularisation term aims to enhance the transferability of learnt features. The establishment of such a transferable relationship ensures the augmented target features receive enough attention during the training of teacher model.



\begin{table*}[t]
\begin{center}
\caption{Results on Camelyon16 with source domains being RCC or NSCLC. The best results are in red underlined, and the second best ones are in blue italic. The subscript in each cell is the standard derivation.}
\begin{tabular}{l|ccc|ccc}
\hline
\multirow{2}{*}{\bf Method} & \multicolumn{3}{c|}{\bf NSCLC $\rightarrow$ Camelyon16} & \multicolumn{3}{c}{\bf RCC $\rightarrow$ Camelyon16}\\
& \textbf{AUC}$\uparrow$ & \textbf{F1}$\uparrow$ & \textbf{Accuracy}$\uparrow$ &\textbf{AUC}$\uparrow$ & \textbf{F1}$\uparrow$ & \textbf{Accuracy}$\uparrow$\\
        \hline \hline
        CLAM \cite{lu_data-efficient_2021} & 0.814 $_{0.010}$ & 0.764 $_{0.032}$ & 0.801 $_{0.027 }$ & 0.814 $_{0.0010}$ & 0.764 $_{0.032}$ & 0.801 $_{0.027}$\\
        Fine-tuning & \sbest{0.885} $_{0.012}$ & \sbest{0.831} $_{0.027}$ & \sbest{0.853} $_{0.021}$ & \sbest{0.872} $_{0.044}$ & \sbest{0.844} $_{0.035}$ & \sbest{0.855} $_{0.024}$\\
        ST~\cite{hinton_distil_2015} & 0.803 $_{0.016 }$ & 0.747 $_{0.017}$ & 0.793 $_{0.012 }$ & 0.820 $_{0.012 }$ & 0.742 $_{0.012 }$ & 0.788 $_{0.009}$\\
        NST~\cite{huang_nst_2017} & 0.824 $_{0.080 }$ & 0.758 $_{0.063 }$ & 0.796 $_{0.043 }$ & 0.814 $_{0.023}$ & 0.742 $_{0.007}$ & 0.788 $_{0.005}$\\
        AT~\cite{zagoruyko_paying_2017} & 0.821 $_{0.011 }$ & 0.760 $_{0.030}$ & 0.804 $_{0.022 }$ & 0.847 $_{0.026}$ & 0.763 $_{0.045}$ & 0.804 $_{0.024 }$\\
        PKT~\cite{passalis_pkt_2018} & 0.794 $_{0.022 }$ & 0.739 $_{0.016}$ & 0.788 $_{0.009}$ & 0.813 $_{0.025}$ & 0.743 $_{0.012}$ & 0.791 $_{0.008}$\\
        CC~\cite{peng_cc_2019} & 0.858 $_{0.054 }$ & 0.801 $_{0.055}$ & 0.827 $_{0.040}$ & 0.839 $_{0.063 }$ & 0.777 $_{0.064 }$ & 0.814 $_{0.040 }$\\
        SP~\cite{tung_sp_2019} & 0.819 $_{0.012 }$ & 0.797 $_{0.023 }$ & 0.827 $_{0.016 }$ & 0.813 $_{0.025 }$ & 0.743 $_{0.012 }$ & 0.791 $_{0.008 }$\\
        PTS~\cite{he_knowledge_2022} & 0.835 $_{0.037 }$ & 0.780 $_{0.031 }$ & 0.811 $_{0.020 }$ & 0.795 $_{0.020 }$ & 0.754 $_{0.006}$ & 0.793 $_{0.005}$\\
        \hline \hline
        \rowcolor{LightCyan}
        TAKT & \best{0.952} $_{0.011}$ & \best{0.896} $_{0.012}$ & \best{0.904} $_{0.012}$ & \best{0.926} $_{0.010 }$ & \best{0.854} $_{0.013 }$ & \best{0.866} $_{0.012 }$\\
        \rowcolor{LightCyan}
        p-value & 0.0006 & 0.0049 & 0.0049 & 0.0027 & 0.1598 & 0.1340\\
        \hline
\end{tabular}
\label{tab:rcc_nsclc_to_camelyon}
\end{center}
\end{table*}

\begin{table*}[t]
\begin{center}
\caption{Results on NSCLC and RCC transferring to each other.}
\begin{tabular}{l|ccc|ccc}
\hline
\multirow{2}{*}{\bf Method} & \multicolumn{3}{c|}{\bf RCC $\rightarrow$ NSCLC} & \multicolumn{3}{c}{\bf NSCLC $\rightarrow$ RCC}\\
& \textbf{AUC}$\uparrow$ & \textbf{F1}$\uparrow$ & \textbf{Accuracy}$\uparrow$ &\textbf{AUC}$\uparrow$ & \textbf{F1}$\uparrow$ & \textbf{Accuracy}$\uparrow$\\
        \hline \hline
        CLAM \cite{lu_data-efficient_2021} & 0.936 $_{0.015}$ & 0.859  $_{0.007  }$ & 0.859  $_{0.007 }$ & 0.970  $_{0.003 }$ & 0.859  $_{0.010}$ & 0.886 $_{0.007 }$\\
        Fine-tuning & 0.944 $_{0.002}$ & 0.876 $_{0.008}$ & 0.876 $_{0.008}$ & 0.979 $_{0.002}$ & \sbest{0.874} $_{0.010 }$ & 0.899 $_{0.009}$\\
        ST~\cite{hinton_distil_2015} & 0.942 $_{0.004}$ & 0.865 $_{0.004}$ & 0.866 $_{0.004}$ & 0.977 $_{0.006}$ & 0.860 $_{0.006}$ & 0.889 $_{0.007}$\\
        NST~\cite{huang_nst_2017} & 0.943 $_{0.007}$ & 0.874 $_{0.007}$ & 0.875 $_{0.007}$ & \sbest{0.980} $_{0.003 }$ & 0.871  $_{0.006  }$ & \sbest{0.900}  $_{0.003  }$\\
        AT~\cite{zagoruyko_paying_2017} & \sbest{0.949} $_{0.001}$ & \sbest{0.877} $_{0.004}$ & \sbest{0.877} $_{0.004}$ & 0.977 $_{0.002 }$ & 0.863 $_{0.003  }$ & 0.893 $_{0.000}$\\
        PKT~\cite{passalis_pkt_2018} & 0.944 $_{0.001  }$ & 0.872 $_{0.012}$ & 0.872 $_{0.012}$ & 0.978  $_{0.005  }$ & 0.865  $_{0.014  }$ & 0.895  $_{0.011  }$\\
        CC~\cite{peng_cc_2019} & 0.945 $_{0.004}$ & 0.872  $_{0.012  }$ & 0.872  $_{0.012  }$ & 0.977 $_{0.005  }$ & 0.864  $_{0.008  }$ & 0.893  $_{0.004  }$\\
        SP~\cite{tung_sp_2019} & 0.937 $_{0.004}$ & \sbest{0.877} $_{0.006}$ & \sbest{0.877} $_{0.006}$ & 0.979 $_{0.001}$ & \sbest{0.874} $_{0.003}$ & 0.896 $_{0.003}$\\
        PTS~\cite{he_knowledge_2022} & 0.942 $_{0.003}$ & 0.864 $_{0.009}$ & 0.864 $_{0.009}$ & 0.970 $_{0.002 }$ & 0.865  $_{0.004}$ & 0.885 $_{0.004}$\\
        \hline \hline
        \rowcolor{LightCyan}
        TAKT & \best{0.955} $_{0.003}$ & \best{0.890} $_{0.010}$ & \best{0.890} $_{0.010}$ & \best{0.984} $_{0.002}$ & \best{0.890} $_{0.008}$ & \best{0.909} $_{0.007}$\\
        \rowcolor{LightCyan}
        p-value & 0.0048 & 0.0288 & 0.0288 & 0.0429 & 0.0072 & 0.0253\\
        \hline
\end{tabular}
\label{tab:rcc_nsclc}
\end{center}
\end{table*}

\section{Experiments and Results}

\subsection{Dataset Descriptions}\label{sec:dataset_desc}
\textbf{Camelyon16.} The Camelyon16 dataset \cite{litjens_1399_2018} contains 399 WSIs of lymph nodes from women with breast cancer. The purpose of this dataset is metastasis detection. The training and test sets contain 270 and 129 WSIs, respectively. We further split the provided training set into training and validation datasets by 8:2 and compare the performances with other methods on the official test dataset. 

\subsubsection{TCGA-RCC (RCC).} RCC contains 940 WSIs, with 121 WSIs from 109 cases of Kidney Chromophobe Renal Cell Carcinoma (KICH), 519 WSIs from 513 cases of Kidney Renal Clear Cell Carcinoma (KIRC), and 300 WSIs from 276 cases of Kidney Renal Papillary Cell Carcinoma (KIRP). The dataset is split into training, validation, and test sets by the ratio of 6:1.5:2.5.

\subsubsection{TCGA-NSCLC (NSCLC).} NSCLC contains 1,053 WSIs, with 512 WSIs from 478 cases of Lung Squamous Cell Carcinoma (LUSC), and 541 WSIs from 478 cases of Lung Adenocarcinoma (LUAD). The dataset split is the same as RCC.

\subsection{Implementation Details}\label{sec:implementation}
\subsubsection{Evaluation Metrics.} \textbf{A}rea \textbf{U}nder the \textbf{C}urve (AUC), F1 and accuracy are the evaluation metrics. These metrics can holistically reflect the overall performances of the models. The thresholds of F1 and accuracy scores are set to 0.5. We also conduct a significance test to further assess the significance of the difference between the means of the highest metrics and those of the second-highest metrics.

\subsubsection{Training Settings.} Methods comparing with ours include no knowledge transfer, fine-tuning, \textbf{S}oft \textbf{T}arget (ST) \cite{hinton_distil_2015}, \textbf{N}euron \textbf{S}electivity \textbf{T}ransfer (NST) \cite{huang_nst_2017}, \textbf{A}ttention \textbf{T}ransfer (AT) \cite{zagoruyko_paying_2017}, \textbf{P}robabilistic \textbf{K}nowledge \textbf{T}ransfer (PKT) \cite{passalis_pkt_2018}, \textbf{C}orrelation \textbf{C}ongruence (CC) \cite{peng_cc_2019}, \textbf{S}imilarity \textbf{P}reserving (SP) \cite{tung_sp_2019} and PTS norm \cite{he_knowledge_2022}. The base model is \textbf{CL}ustering-constrained-\textbf{A}ttention \textbf{M}ultiple-instance learning (CLAM) \cite{lu_data-efficient_2021}. The student model in our method is initialised with the teacher model. The model is trained up to 200 epochs and no less than 50 epochs. The training is ceased when the validation loss stops decreasing for 20 epochs. All experiments are repeated three times with different random seeds. The means and standard deviations of the performances are reported.

\subsubsection{Hyper-parameters.} The learning rate, weight decay and dropout are set to 2$\times$10$^{-4}$, 1$\times$10$^{-5}$, and 0.25, respectively \cite{lu_data-efficient_2021}. Adam optimiser is used \cite{lu_data-efficient_2021}. The probability of augmenting each feature $p$ and the strength of augmentation $\lambda$ is set to 0.3 and 0.5, respectively \cite{yang_remix_2022}. The number of MHA heads is 8. WSIs are split into non-overlapping patches of 256 $\times$ 256 pixels at 20$\times$ magnification \cite{lu_data-efficient_2021}. ResNet-50 \cite{resnet_he_16} pre-trained on ImageNet is used to extract features from them. $r, r_1, r_2$ are set to 0.1, 0.5, 0.5, respectively \cite{xu_2023_motcat}. We perform sensitivity analysis on $\alpha, \beta$ in supplementary materials, and they are set to 0.1, 0.2, respectively.

\subsection{Comparison Results}
We compare our method with other related knowledge transfer methods in four settings: NSCLC to Camelyon16, RCC to Camelyon16, RCC to NSCLC and NSCLC to RCC. The experimental results are reported in \cref{tab:rcc_nsclc_to_camelyon} and \cref{tab:rcc_nsclc}. Our method achieves the best performance across every metric, especially on Camelyon16, where our method outperforms other methods by a large margin. In addition, the p-values comparing the best (ours) and second-best metrics indicate that our method significantly outperforms second-best performing methods. Furthermore, due to the small tumour size and the limited number of samples, Camelyon16 is more difficult than TCGA datasets, which is reflected in the absolute value of the metrics. The results prove that our method can effectively transfer knowledge from a simpler dataset to a harder one. Another observation is that methods with additional supervision signals perform better on TCGA datasets, and fine-tuning performs better when transferring knowledge from TCGA to the Camelyon16 dataset. The potential reason for this is that TCGA datasets are more similar, as evidenced by the maximum mean discrepancy scores between these datasets. The tumour size is drastically different between Camelyon16 and TCGA datasets, leading to significant differences in attention distributions. Methods with additional supervision signals (features, attention scores and logits) all rely on the attention scores as bag features are eventually derived from them. Therefore, these methods may consistently introduce bias during student training, resulting in lower performance. 

\begin{table*}[t]
\begin{center}
\caption{Ablation studies on NSCLC to Camelyon16 transfer. Left: Ablation study results on our ``replace'' augmentation method (R), MSE, cosine similarity (cos), and OT distance with dimension reduction. Right: The augmentation methods are denoted as: Append (A), Replace (R), Interpolate (I), Covary (C), and Joint (J).}
\begin{tabular}{l|ccc||l|ccc}
\hline
\multirow{2}{*}{\bf Method} & \multicolumn{3}{c||}{\bf NSCLC $\rightarrow$ Camelyon16} & \multirow{2}{*}{\bf Method} & \multicolumn{3}{c}{\bf NSCLC $\rightarrow$ Camelyon16}\\
& \textbf{AUC}$\uparrow$ & \textbf{F1}$\uparrow$ & \textbf{Acc.}$\uparrow$ & & \textbf{AUC}$\uparrow$ & \textbf{F1}$\uparrow$ & \textbf{Acc.}$\uparrow$\\
        \hline \hline
        No Aug. & 0.921$_{0.019}$ & 0.855$_{0.046 }$ & 0.871$_{0.038 }$ & +A,OT & 0.939$_{0.018}$ & \sbest{0.885}$_{0.031}$ & \sbest{0.897}$_{0.025}$ \\
        +R & 0.932$_{0.013}$ & 0.862$_{0.016}$ & 0.873$_{0.016}$ & +R,OT & \best{0.952}$_{0.011}$ &\best{0.896}$_{0.012}$ & \best{0.904}$_{0.012}$\\
        +R,MSE & 0.942$_{0.007}$ & 0.861$_{0.012}$ & 0.873$_{0.012}$ & +I,OT & 0.933$_{0.021 }$ & 0.854$_{0.038 }$ & 0.871$_{0.031 }$ \\
        +R,cos & \sbest{0.943}$_{0.004}$ & \sbest{0.877}$_{0.006}$ & \sbest{0.889}$_{0.005}$ & +C,OT & 0.931$_{0.012 }$ & 0.854$_{0.018}$ &0.868$_{0.013}$\\
        +R,OT & \best{0.952}$_{0.011}$ &\best{0.896}$_{0.012}$ & \best{0.904}$_{0.012}$ & +J,OT & \sbest{0.941}$_{0.004 }$ & 0.865$_{0.013}$ &0.876$_{0.013}$\\
        \hline
\end{tabular}
\label{tab:ablation_studies}
\end{center}
\end{table*}

\subsection{Ablation Studies}

\subsubsection{TAFA Module.} We investigate the impact of ``replace'' augmentation with different alignment losses on the final performance, including \textbf{M}ean \textbf{S}quared \textbf{E}rror (MSE), cosine similarity (cos) and OT distance. The experimental results are shown in the left part of \cref{tab:ablation_studies}. When we incorporate the ``replace'' augmentation during teacher model training, the AUC score increases by 1.10\%. With MSE, cos and OT, the AUC score further increases by 1.00\% and 1.10\% and 2.00\%, respectively, demonstrating the effectiveness of OT distance.

\subsubsection{Augmentation Methods.} We conduct experiments using the aforementioned augmentation methods. The experimental results are presented in the right part of \cref{tab:ablation_studies}. Incorporating augmentation methods results in improved AUC scores compared to the absence of augmentation, with a minimum increase of 1.00\%, highlighting their effectiveness. Particularly, the ``replace'' augmentation method achieved the highest performance, surpassing other methods by a significant margin (with a best average AUC score of 95.2\%).

\section{Conclusion}
In this work, we presented a TAKT framework specifically designed for WSI classification. The framework contains two main components: the TADA method and the TAFA module. The TADA method augments the source dataset by actively retrieving the closest centroids from the target domain, facilitating the teacher model to acquire common knowledge from both domains. To mitigate the bias of teacher model towards the source domain, we proposed a TAFA module to establish a latent relationship between source and target features by solving a OT problem, enforcing the teacher model to pay similar attention to features from both the source and target domains. 
Our experimental results demonstrated that models trained with knowledge transfer techniques outperformed those trained from scratch and our method achieved state-of-the-art performance among other adapted knowledge transfer methods.

\subsubsection{\ackname} The research presented in this paper was partially supported by the Research Grants Council of the Hong Kong Special Administrative Region, China (CUHK 14222922, RGC GRF 2151185).
\subsubsection{\discintname}
The authors have no competing interests to declare that are relevant to the content of this article.
%
%
%
\bibliographystyle{splncs04}
\bibliography{main}

\begin{thebibliography}{10}
\providecommand{\url}[1]{\texttt{#1}}
\providecommand{\urlprefix}{URL }
\providecommand{\doi}[1]{https://doi.org/#1}

\bibitem{ahn_2020_uda}
Ahn, E., Kumar, A., Fulham, M., Feng, D., Kim, J.: Unsupervised domain adaptation to classify medical images using zero-bias convolutional auto-encoders and context-based feature augmentation. IEEE Transactions on Medical Imaging  \textbf{39}(7),  2385--2394 (2020)

\bibitem{amupan_predictors_2021}
Aumpan, N., Vilaichone, R.k., Pornthisarn, B., Chonprasertsuk, S., Siramolpiwat, S., Bhanthumkomol, P., Nunanan, P., Issariyakulkarn, N., Ratana-Amornpin, S., Miftahussurur, M., et~al.: Predictors for regression and progression of intestinal metaplasia (im): a large population-based study from low prevalence area of gastric cancer (im-predictor trial). PloS One  \textbf{16}(8),  e0255601 (2021)

\bibitem{baba2007comparative}
Baba, A.I., C{\^a}toi, C.: Comparative oncology. Publishing House of the Romanian Academy (2007)

\bibitem{campanella_clinical-grade_2019}
Campanella, G., Hanna, M.G., Geneslaw, L., Miraflor, A., Werneck Krauss~Silva, V., Busam, K.J., Brogi, E., Reuter, V.E., Klimstra, D.S., Fuchs, T.J.: Clinical-grade computational pathology using weakly supervised deep learning on whole slide images. Nature Medicine  \textbf{25}(8),  1301--1309 (2019)

\bibitem{chen_annotation-free_2021}
Chen, C.L., Chen, C.C., Yu, W.H., Chen, S.H., Chang, Y.C., Hsu, T.I., Hsiao, M., Yeh, C.Y., Chen, C.Y.: An annotation-free whole-slide training approach to pathological classification of lung cancer types using deep learning. Nature Communications  \textbf{12}(1), ~1193 (2021)

\bibitem{feng2021deep}
Feng, Y., Xu, X., Wang, Y., Lei, X., Teo, S.K., Sim, J.Z.T., Ting, Y., Zhen, L., Zhou, J.T., Liu, Y., et~al.: Deep supervised domain adaptation for pneumonia diagnosis from chest x-ray images. IEEE Journal of Biomedical and Health Informatics  \textbf{26}(3),  1080--1090 (2021)

\bibitem{frogner_learning_2015}
Frogner, C., Zhang, C., Mobahi, H., Araya{-}Polo, M., Poggio, T.A.: Learning with a wasserstein loss. In: Advances in Neural Information Processing Systems. pp. 2053--2061 (2015)

\bibitem{resnet_he_16}
He, K., Zhang, X., Ren, S., Sun, J.: Deep residual learning for image recognition. In: IEEE/CVF Conference on Computer Vision and Pattern Recognition. pp. 770--778 (2016)

\bibitem{he_knowledge_2022}
He, R., Sun, S., Yang, J., Bai, S., Qi, X.: Knowledge distillation as efficient pre-training: Faster convergence, higher data-efficiency, and better transferability. In: IEEE/CVF Conference on Computer Vision and Pattern Recognition. pp. 9161--9171 (2022)

\bibitem{hinton_distil_2015}
Hinton, G., Vinyals, O., Dean, J.: Distilling the knowledge in a neural network. arXiv preprint arXiv:1503.02531  (2015)

\bibitem{huang_nst_2017}
Huang, Z., Wang, N.: Like what you like: Knowledge distill via neuron selectivity transfer. arXiv preprint arXiv:1707.01219  (2017)

\bibitem{ianni_tailored_2020}
Ianni, J.D., Soans, R.E., Sankarapandian, S., Chamarthi, R.V., Ayyagari, D., Olsen, T.G., Bonham, M.J., Stavish, C.C., Motaparthi, K., Cockerell, C.J., et~al.: Tailored for real-world: a whole slide image classification system validated on uncurated multi-site data emulating the prospective pathology workload. Scientific Reports  \textbf{10}(1), ~3217 (2020)

\bibitem{ilse_attention-based_2018}
Ilse, M., Tomczak, J.M., Welling, M.: Attention-based deep multiple instance learning. In: International Conference on Machine Learning. Proceedings of machine learning research, vol.~80, pp. 2132--2141. {PMLR} (2018)

\bibitem{keikhosravi_non-disruptive_2020}
Keikhosravi, A., Li, B., Liu, Y., Conklin, M.W., Loeffler, A.G., Eliceiri, K.W.: Non-disruptive collagen characterization in clinical histopathology using cross-modality image synthesis. Communications Biology  \textbf{3}(1), ~414 (2020)

\bibitem{litjens_1399_2018}
Litjens, G., Bandi, P., Ehteshami~Bejnordi, B., Geessink, O., Balkenhol, M., Bult, P., Halilovic, A., Hermsen, M., van~de Loo, R., Vogels, R., et~al.: 1399 h\&e-stained sentinel lymph node sections of breast cancer patients: the camelyon dataset. GigaScience  \textbf{7}(6),  giy065 (2018)

\bibitem{litjens_deep_2016}
Litjens, G., S{\'a}nchez, C.I., Timofeeva, N., Hermsen, M., Nagtegaal, I., Kovacs, I., Hulsbergen-Van De~Kaa, C., Bult, P., Van~Ginneken, B., Van Der~Laak, J.: Deep learning as a tool for increased accuracy and efficiency of histopathological diagnosis. Scientific Reports  \textbf{6}(1),  26286 (2016)

\bibitem{lloyd_kmeans_1982}
Lloyd, S.: Least squares quantization in pcm. IEEE Transactions on Information Theory  \textbf{28}(2),  129--137 (1982)

\bibitem{lu_data-efficient_2021}
Lu, M.Y., Williamson, D.F., Chen, T.Y., Chen, R.J., Barbieri, M., Mahmood, F.: Data-efficient and weakly supervised computational pathology on whole-slide images. Nature Biomedical Engineering  \textbf{5}(6),  555--570 (2021)

\bibitem{pan_survey_2010}
Pan, S.J., Yang, Q.: A survey on transfer learning. IEEE Transactions on Knowledge and Data Engineering  \textbf{22}(10),  1345--1359 (2009)

\bibitem{passalis_pkt_2018}
Passalis, N., Tefas, A.: Learning deep representations with probabilistic knowledge transfer. In: European Conference on Computer Vision. pp. 268--284 (2018)

\bibitem{peng_cc_2019}
Peng, B., Jin, X., Liu, J., Li, D., Wu, Y., Liu, Y., Zhou, S., Zhang, Z.: Correlation congruence for knowledge distillation. In: IEEE/CVF International Conference on Computer Vision. pp. 5007--5016 (2019)

\bibitem{shao_transmil}
Shao, Z., Bian, H., Chen, Y., Wang, Y., Zhang, J., Ji, X., Zhang, Y.: Transmil: Transformer based correlated multiple instance learning for whole slide image classification. In: Advances in Neural Information Processing Systems. pp. 2136--2147 (2021)

\bibitem{tung_sp_2019}
Tung, F., Mori, G.: Similarity-preserving knowledge distillation. In: IEEE/CVF International Conference on Computer Vision. pp. 1365--1374 (2019)

\bibitem{transformer_2017}
Vaswani, A., Shazeer, N., Parmar, N., Uszkoreit, J., Jones, L., Gomez, A.N., Kaiser, L., Polosukhin, I.: Attention is all you need. In: Advances in Neural Information Processing Systems. pp. 5998--6008 (2017)

\bibitem{xiong_diagnose_2023}
Xiong, C., Chen, H., Sung, J.J., King, I.: Diagnose like a pathologist: Transformer-enabled hierarchical attention-guided multiple instance learning for whole slide image classification. In: International Joint Conference on Artificial Intelligence. pp. 1587--1595 (2023)

\bibitem{xu_2023_motcat}
Xu, Y., Chen, H.: Multimodal optimal transport-based co-attention transformer with global structure consistency for survival prediction. In: IEEE/CVF International Conference on Computer Vision. pp. 21241--21251 (October 2023)

\bibitem{yang_remix_2022}
Yang, J., Chen, H., Zhao, Y., Yang, F., Zhang, Y., He, L., Yao, J.: Remix: A general and efficient framework for multiple instance learning based whole slide image classification. In: International Conference on Medical Image Computing and Computer-Assisted Intervention. pp. 35--45. Springer (2022)

\bibitem{yu_transfer_2022}
Yu, X., Wang, J., Hong, Q.Q., Teku, R., Wang, S.H., Zhang, Y.D.: Transfer learning for medical images analyses: A survey. Neurocomputing  \textbf{489},  230--254 (2022)

\bibitem{zagoruyko_paying_2017}
Zagoruyko, S., Komodakis, N.: Paying more attention to attention: Improving the performance of convolutional neural networks via attention transfer. In: International Conference on Machine Learning (2017)

\bibitem{zhuang_comprehensive_2021}
Zhuang, F., Qi, Z., Duan, K., Xi, D., Zhu, Y., Zhu, H., Xiong, H., He, Q.: A comprehensive survey on transfer learning. Proceedings of the IEEE  \textbf{109}(1),  43--76 (2020)

\end{thebibliography}

\end{document}